\documentclass[cameraready]{Interspeech}
\usepackage{makecell}

\title{Listenability of Synthetic Speech: On the Effect of Linguistic Registers in Text-to-Speech Input}

\author[affiliation={1,2}, orcid=0009-0003-7489-4131]{Maja Jønck}{Hjuler}
\author[affiliation={1}]{Tuyet Katie}{Nhi Tran}
\author[affiliation={1}, orcid=0000-0003-2359-2515]{Laurianne}{Sitbon}

\address{
    $^1$School of Computer Science, Queensland University of Technology, Brisbane, Australia \\
    $^2$University Grenoble Alpes, CNRS, Grenoble INP, LIG, 38000 Grenoble, France 
}

\email{majajoenck.hjuler@hdr.qut.edu.au, n11159243@qut.edu.au, l.sitbon@qut.edu.au}

\keywords{Listenability, Text-to-Speech Synthesis, Human-Computer Interaction, Spoken Language Comprehension}

\usepackage{comment}

\begin{document}

\maketitle

\begin{abstract}

With the rapid advancement of Large Language Models (LLMs), LLM-generated text is increasingly being used as input for speech synthesis, yet little is known about its listenability compared to other linguistic registers. 
This paper presents a preliminary study of listenability across conversational, formal written, simplified written, and LLM-generated registers in terms of listener effort and comprehension during synthesized delivery.
An experiment with 47 participants measured subjective mental effort, cognitive load, and cued recall in a within-subjects design. Register characteristics were examined through readability metrics and Biber's Multi-Dimensional Analysis. 
Our findings indicate higher listenability for synthesized speech from written sources compared to radio transcripts, demonstrating the intricacies of language and register. 
Furthermore, LLM-generated text was as listenable as Wikipedia-sourced text, supporting its suitability in synthetic speech applications.


\end{abstract}

\section{Introduction}

Advances in AI agents have advanced communication with machines, enabling a new range of interaction modalities and realizing the potential of new forms of human-computer interaction.
Humanlike communication is crucial for conversational agents \cite{Kim2025}, and is considered more important than humanlike behavior and appearance in robot companions \cite{Dautenhahn_2005}.
Meanwhile, speech is one of our primary means of communication, and there has been a rise in the popularity of virtual assistants that offer voice-centered interaction \cite{Cho2019-bh}.
One way to assess synthesized speech is through 'listenability', or ‘listening effort’, which is the mental and/or cognitive effort required to process spoken language~\cite{Winn2021}. Most conversational Text-to-Speech (TTS) pipelines are driven by an upstream text generator, e.g.~\cite{Kim2025, Wang2024, Zhou2024}, and modern TTS systems are optimized for Large Language Model (LLM) output \cite{Casanova2024}.
This assumes that LLM output is suitable for speech generation. 
Yet, few studies have considered the suitability of written sources and LLM-generated text for audio delivery, leaving a gap in our understanding of how the linguistic properties of TTS input affect the listening experience.

Language varies according to the register, functional context, or communicative situation in which the language is intended \cite{Biber_2019}. 
For example, a conversation with a close friend differs greatly from the language used in a formal academic paper.
Biber’s Multi-Dimensional Analysis (MDA) \cite{Biber_1988} provides a framework for quantifying this variation, extracting six functional dimensions from patterns of co-occurring linguistic features in spoken and written English.
The most prominent, Dimension~1, ranges from involved, interactional speech to informationally dense, formal prose (Table \ref{tab:MDA_dimensions}).
Since LLMs are predominantly trained on written text \cite{Cho2024}, the registers they produce may not align with the demands of spoken delivery.
In general, the role of the TTS input text’s linguistic register in driving listenability remains largely unexplored.


This preliminary study aims to highlight this research gap and further the understanding of listenability by investigating the influence of language variation in TTS input.
Evaluations of synthetic speech typically focus on \textit{naturalness}, rated via mean opinion scores (MOS), or \textit{intelligibility}, measured by word error rate.
However, these measures assess surface-level perceptual quality, not the ease with which a listener can process and retain spoken content.
To support the usability of TTS systems in real-world applications, we conduct subjective evaluations with human listeners to assess TTS performance.
We investigate the following \textit{research questions}: 
How does the listenability of speech synthesized from written registers compare to that of synthesized conversational speech?
And: Do register differences across written sources translate into measurable differences in listenability when delivered via speech synthesis?
We find that speech generated from speech transcripts is the least listenable compared to Wikipedia-sourced and LLM-generated text ($p<0.001$), with no statistically significant difference between the written sources.\footnote{\label{data_release}Stimuli and analysis files are available at \url{https://github.com/MajaHjuler/TTS_Listenability}}

\section{Related Work}


TTS research has primarily focused on the perceptual quality of synthesized speech output, i.e., clarity, intonation, stress, and pronunciation, evaluated through MOS and intelligibility \cite{Cooper2024}.
While these metrics are well-established, they assess surface-level acoustic quality of the output and do not necessarily capture the cognitive effort required to process and retain spoken content~\cite{Winn2021}.
Govender et al.~\cite{Govender2023} investigated the cognitive load of listening to synthetic speech using the dual-task paradigm and pupillometry, demonstrating that synthesizer quality affects listening effort.
However, their focus was on differences between synthesizers, i.e., output quality, rather than on the linguistic properties of the TTS input text.

Text generation and simplification models have largely optimized their output for readability, defined as the ease of comprehending written text~\cite{Agraval_2024, Sastre_2024}.
Cho et al.~\cite{Cho2024} explored methods to adapt LLM-generated text to the speech domain, including specific prompting strategies. 
They used automated 'speech-worthiness' metrics for the input text, such as Flesch Reading Ease, sentence length, and syntactic dependency depth, some of which are readability metrics.
However, as argued by Rubin~\cite{Rubin01102012}, listenability is distinct from readability and must be evaluated independently.
Factors enhancing reading do not necessarily ease listening, as spoken and written language impose different processing demands~\cite{Rubin01042000, Rubin01102012}.
In assistive technologies, accessibility guidelines similarly emphasize this distinction~\cite{Machuca-Ayuso2025-uk}.
Woo et al.~\cite{woo-etal-2025-think} noted the mismatch between optimal textual and spoken delivery, proposing an intermediate verbalizing step to convert LLM output into speech-ready TTS input.
They perform human evaluations based on a 5-point Likert scale across four criteria: naturalness, conciseness, understandability, and overall quality~\cite{woo-etal-2025-think}.
Meanwhile, Govender et al.~\cite{Govender2023} argue that reducing cognitive load in listening is crucial for the user experience of TTS systems.
Yet, perceptual experiments targeting listenability remain scarce.

Language varies systematically across communicative contexts, or registers~\cite{Biber_2019}, and this variation affects spoken delivery.
Biber's MDA has been widely applied to characterize register variation in English corpora along functional dimensions for informational density and narrative stance~\cite{Biber_1988, Nini2019}.
Since LLMs are trained predominantly on written sources~\cite{Cho2024}, their output occupies a different register position than naturally spoken language.
Differences in perceived naturalness between spontaneous speech and read-aloud text have been documented~\cite{Dall_2014}, and linguistic differences between human and LLM-generated dialogue have been identified~\cite{sandler2024linguisticcomparisonhumanchatgptgenerated}, showing greater variability and authenticity in human dialogues.
However, linguistic variation analysis has not been applied to characterize TTS input sources, nor has register position been linked to listenability outcomes in synthesized speech - a gap this study addresses.

\begin{table}[t!]
\caption{Descriptions of the six MDA dimensions with examples (non-exhaustive) of the corresponding linguistic features \cite{Biber_1988}. }
\label{tab:MDA_dimensions}
\centering
\begin{tabular}{l}
\hline

\textbf{D1. Involved vs.~informational}\\ \hline
\textit{High: affective/interactional. Low: informationally dense.} 
\\ 
\makecell[l]{Present tenses, 1st and 2nd person pronouns, nouns, \\ average word length, prepositions, attributive adjectives.}\\
\hline

\textbf{D2. Narrative vs.~non-narrative}\\ \hline
\textit{High: narrative text.} \\
\makecell[l]{Past tenses, 3rd person pronouns, perfect aspects, public \\ verbs, synthetic negations, present participial clauses.}\\
\hline

\textbf{D3. Explicit vs.~situation-dependent}\\ \hline
\textit{High: context-independent.}\\
\makecell[l]{Wh-relative clauses on object or subject position, adverbs, \\ nominalisations, time and place adverbials.}\\
\hline

\textbf{D4. Overt persuasion}\\ \hline
\textit{High: marks point of view and assessment of certainty.}
\\
\makecell[l]{Infinitives, prediction modals, suasive verbs, conditional \\ subordinations, necessity modals.}\\
\hline

\textbf{D5. Abstract vs.~non-abstract}\\ \hline
\textit{High: technical, abstract and formal.}
\\
\makecell[l]{Conjuncts, agentless passives, past participial clauses, \\ WHIZ deletion relatives, other adverbial subordinators.}\\
\hline

\textbf{D6. Online informational elaboration}\\ \hline
\textit{High: informational but time-constrained.}
\\
\makecell[l]{That clauses as verb and adjective complements, demon- \\ stratives, that relative clauses on object position.}\\
\hline
\vspace{-0.8cm}
\end{tabular}
\end{table}

\section{Methods}

In an online Qualtrics survey, participants were instructed to listen to a series of auditory stimuli presented in different registers.  
The study employed a repeated-measures experimental design, utilising self-report measures of mental effort and cued recall memory questions targeting the word level.
There was an institutional ethical clearance for this study \textbf{(approval \#[anon.]} from Queensland University of Technology).

\subsection{Procedure}
Data collected through the online survey included basic demographic information, including age, gender, and whether English was the participant’s first language. 
Participants were briefed on the objectives of the research project. 
They then listened to an audio sample and answered corresponding mental effort and cued recall questions. 
This process was repeated for a total of 16 audio samples; four samples for each of four types of auditory stimuli generated from text. 
Each sample covered a different topic.
The stimuli were presented to the participants in randomised order for both the type and the topic. 
Mental effort questions were used to subjectively assess the cognitive load associated with listening. 
These included a 5-item NASA Task Load Index (NASA TLX) questionnaire \cite{HART1988139}, and a 3-item Amount of Invested Mental Effort (AIME) questionnaire \cite{Rubin01102012, Rubin01042000}, which measures the subjective effort
a listener invests in comprehending spoken language.
One of the six items from the original NASA TLX questionnaire was removed as the item referred to physical demand and did not apply to the study.
Similarly, one of the four items of the AIME questionnaire was removed as it referred to a wider audience.
All measures were recorded on a continuous scale from 1 to 100 using a slider interface in Qualtrics.
Cued recall multiple-choice questions were developed based on sample topics using a word-level approach comprehension framework \cite{Chapman2020}, with three recall questions concerning the theme and contents.
The order of the multiple-choice answers to the questions was randomized.

\subsection{Participants}

The 47 participants were aged 18 to 50 (\textit{M} = 24.7, SD = 5.9). 
26 (55.3\%) of the participants were female, 19 (40.4\%) were male, and two (4.3\%) preferred not to disclose their gender. 
30 (63.8\%) of the participants identified English as their first language. 
Participants were recruited through personal networks using social media such as Facebook, Discord, Instagram, Microsoft Teams, and Slack, as well as physical posters with hyperlinks and QR codes around the university campus.

\subsection{Stimuli selection}
Four stimulus types were chosen, which occupy distinct positions on Biber's register continuum \cite{Biber_1988}. 
The four stimulus types were:
\textbf{ART}: Transcribed multi-party talkback radio recordings (host and callers) from the Australian Radio Talkback (ART) Corpus \cite{Smith2023}.
\textbf{WSiE}: Written formal simple language sample using the Wikipedia Simple English
API\footnote{\url{https://simple.wikipedia.org/api/rest_v1/}}.
\textbf{GPT}: LLM response using OpenAI's gpt-4o-mini API\footnote{\url{https://developers.openai.com/api/docs/models/gpt-4o-mini}}.
First, four ART Corpus samples were randomly selected and manually sampled.
Wikipedia samples were then collected based on the topics of the ART Corpus samples to ensure consistency.
We sample multiple topics per source to reduce the risk of results being driven by one unusual text.
The final topic list comprised: \textit{University}, \textit{Hubble Ultra-Deep Field}, \textit{Mandarin Trees}, and \textit{Cramp}. 
GPT-4o mini samples for each topic were generated using OpenAI’s API using the prompt \textit{“This is a conversation between people. 
Be as listenable as possible. 
What is \{topic\}”}. 
Minimal prompt engineering was employed in formulating the prompt. 

\begin{table}[h!]
\caption{Automated readability evaluation of each sample (\#1-4) for TTS  text sources (M(SD)). 
W: Word count, SL: Sentence Length, FRE: Flesch Reading Ease, FKGL: Flesch-Kincaid Grade Level Readability, DD: Dependency Depth.}
\label{tab:readability_text_stimuli}
\begin{tabular}{lccccc}
\hline
\multicolumn{1}{c}{\textbf{Stimuli}} & \textbf{W} & \textbf{SL $\downarrow$} & \textbf{FRE $\uparrow$} & \textbf{FKGL $\downarrow$} & \textbf{DD $\downarrow$} \\ \hline
WSiE                                 & \makecell{54.8 \\ (18.7)}       & \makecell{13.1 \\ (3.6)}                    & \makecell{46.7 \\ (16.9)}                   & \makecell{10.0 \\ (1.8)}                      & \makecell{4.4 \\ (1.4)}                     \\
\hline
WStE                                 & \makecell{59.0 \\ (14.8)}      & \makecell{19.7  \\ (4.9)}                    & \makecell{27.2  \\ (23.7)}                   & \makecell{14.4  \\ (2.8)}                      & \makecell{5.9  \\ (0.6)}                     \\ \hline
GPT                                  & \makecell{74.5 \\ (9.8)}      & \makecell{16.9  \\(2.1)}                    & \makecell{40.1  \\ (22.2)}                   & \makecell{11.9  \\ (3.0)}                      & \makecell{6.2  \\ (1.1)}                     \\
                                     \hline
ART & \makecell{154.8 \\ (20.5)} & \makecell{40.7 \\ (19.6)} & \makecell{- \\ } & \makecell{- \\ } & \makecell{6.1 \\ (1.8)}  \\ \hline
\end{tabular}
\vspace{-0.2cm}
\end{table}

All stimuli were generated using Google’s free-tier TTS API\footnote{\url{https://ai.google.dev/gemini-api/docs/speech-generation}}. 
An Australian accent was selected to match the demographic of the majority of the participants, and the same male voice setting was used for all samples. 
Multiple speakers in the ART samples were rendered using additional Australian female and male voice settings. 
Minimal manual adjustments were made to the samples (e.g., insertion of pauses) to enhance sample quality, particularly for the ART samples, as the TTS converter had difficulty with run-on sentences.

\subsection{Analysis}

We analysed the text sources of the generated speech stimuli using automated measures of readability (Table \ref{tab:readability_text_stimuli}).
Flesch Reading Ease (FRE) scores rate text on a 100-point scale, where scores correlate positively with readability \cite{Flesch1948}.
Flesch-Kincaid Grade Level (FKGL) scores rate text at its supposed U.S. school grade level \cite{kincaid1975derivation}.
Although FRE and FKGL are both based on word and sentence length, they have different weighting factors.
Inspired by Cho et al.~\cite{Cho2024}, we also quantified sentence complexity using SpaCy’s dependency parser\footnote{\url{https://spacy.io/api/dependencyparser}}, i.e., the depth of the resulting dependency graph (DD).
To characterize the linguistic register of all auditory stimuli, we applied Biber's MDA framework using the Multidimensional Analysis Tagger (MAT)\footnote{\url{https://sites.google.com/site/multidimensionaltagger/}} \cite{Nini2019}, which quantifies register variation along six functional dimensions defined by 67 co-occurring grammatical and lexical features (e.g., in Table \ref{tab:biber_dimensional_analysis}).
Unlike neural embeddings, which capture register variation implicitly, MDA provides interpretable, feature-level explanations of why texts differ linguistically.
MAT also offers the option to apply a z-score correction, i.e., reduce all z-scores exceeding 5 to 5, to avoid overly inflated z-scores and dimension scores \cite{Nini2019}.
The most similar text types were assigned using Euclidean distance from the centroids of the clusters reported in Biber \cite{biber1989drift, Nini2019}.

Qualtrics responses were cleaned by removing partial submissions. 
A one-way repeated-measures ANOVA was conducted with stimulus type (ART, WSiE, WStE, GPT) as a within-subjects factor ($N = 47$), with separate analyses performed for AIME, NASA TLX, and cued recall. 
Sphericity was assessed using Mauchly’s test, and violations were corrected using the Greenhouse–Geisser adjustment. 
Post-hoc pairwise comparisons were Bonferroni-corrected to control for multiple comparisons. Statistical significance was set at $\alpha = 0.05$.

\section{Results}

\textbf{Readability analysis of TTS input.}
Table~\ref{tab:readability_text_stimuli} presents readability and linguistic complexity evaluations for all four sources.
For ART, FRE and FKGL are omitted as they do not apply to transcribed speech. 
The remaining measures show substantially longer words and sentences, consistent with spontaneous spoken dialogue.
WSiE texts were the shortest and most syntactically simple, averaging the fewest words per sentence and achieving the highest mean FRE score, indicating the greatest ease of reading. 
FKGL scores placed WSiE texts at approximately grade 10, the lowest of all written sources, with low variability suggesting consistency across topics.
Conversely, WStE texts were consistently the most complex, averaging longer sentences, the lowest FRE scores, and an FKGL of approximately grade 14, indicating an undergraduate reading level.
However, a large standard deviation in FRE scores reflects considerable variation across the four topics.
GPT texts fell in between on all metrics, with moderate sentence length, FRE, and FKGL. 
DD is the smallest for WSiE on average, but shows no clear trend for WStE and GPT, indicating similar sentence complexity despite varying lexical difficulty.


\begin{figure}[!t]
    \centering
    \includegraphics[width=\linewidth]{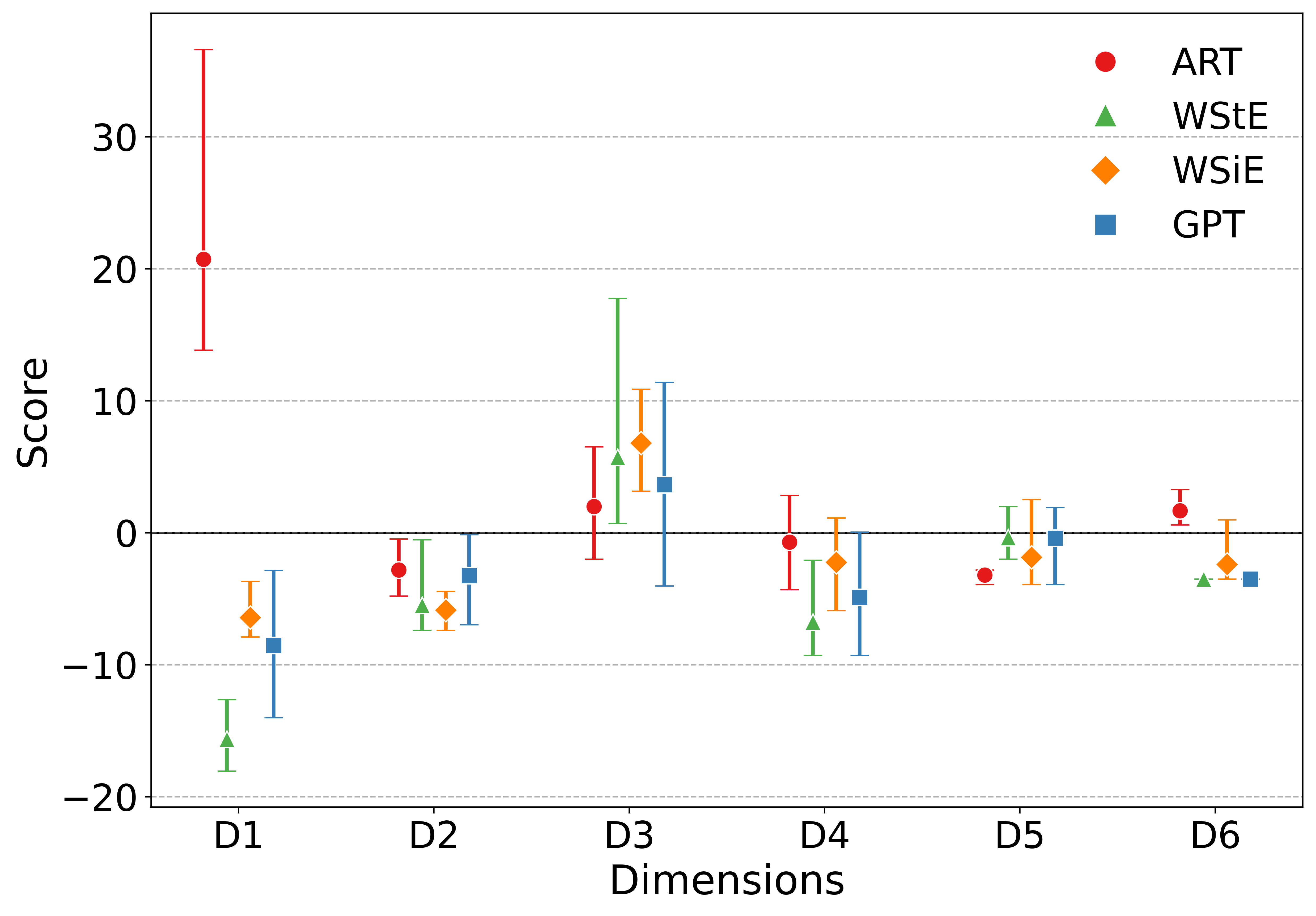}
    \vspace{-0.2cm}
    \caption{MDA scores across all six dimensions (D) for the four stimulus types, computed with z-scores and all linguistic features. 
    Points mark the mean score per source, and error bars indicate within-dataset variability across topics ($\pm$SD).}
    \label{fig:dimensions}
    \vspace{-0.2cm}
\end{figure}

\begin{table}[!t]
    \caption{Closest text types for TTS input as classified by MAT.}
    \centering
    \label{tab:biber_dimensional_analysis}
    \begin{tabular}{llccccl}
                                         \hline
    \textbf{Stimuli}                     & \multicolumn{6}{l}{\textbf{Closest text type(s)}}                                                                                                       \\ \hline
    WSiE                              & \multicolumn{6}{l}{General narrative exposition}                                                                                                               \\
    WStE                              & \multicolumn{6}{l}{Learned exposition, Scientific exposition}                                                                                                              \\
    GPT                                  & \multicolumn{6}{l}{\makecell[l]{General narrative exposition, Involved persuasion,\\ Learned exposition}}                                                                                                     \\
    ART                                  & \multicolumn{6}{l}{Informational interaction, Involved persuasion}                                                                                                        \\
    \hline
    \end{tabular}
    \vspace{-0.2cm}
\end{table}

\noindent\textbf{Stimuli analysis using MAT.}
The MDA analysis (Figure \ref{fig:dimensions} and Table~\ref{tab:biber_dimensional_analysis}) indicates that the four sources are positioned distinctly along the register continuum. 
The ART corpus scores highest on Dimension 1 (D1): Involved vs.~informational, reflecting the conversational nature.
Overall, the closest \textit{text types} are Informational interaction and Involved persuasion.
At the opposite extreme, WStE scores strongly negatively on D1, indicating that it comprises strongly written, formal, informational texts with almost no conversational features.
Compared to ART, WStE contains longer words, technical vocabulary, and no speaker presence. 
The closest text types are Learned and Scientific exposition.
GPT and WSiE occupy an intermediate register, both written and informal, mainly classified as General narrative exposition.
The WSiE samples are placed between ART and WStE on D1, making them more informational than ART but less formal than WStE.
GPT overlaps with WSiE on most dimensions.
Both are informational, noun-heavy registers. 
Overall, these samples are classified as General narrative exposition. 
Across all sources, D2 (Narrative vs.~non-Narrative) and D4 (Overt persuasion) remain consistently negative, confirming the expository, non-argumentative nature of all texts.
For the Wikipedia corpora, this corresponds to the factual nature of encyclopedia articles.

\noindent\textbf{Cognitive effort and listening comprehension.}
As shown in Table \ref{tab:perceptual}, ART imposed the highest cognitive cost as measured by AIME, requiring more than twice the effort of the other sources (\textit{M} = 62.9 vs.~25–29).
GPT required the least effort (\textit{M} = 25.2), with WStE and WSiE closely matched.
A similar pattern was observed for NASA TLX: ART was a clear outlier (\textit{M} = 46.8), whereas GPT, WSiE, and WStE clustered between 14 and 18.
For cued recall, the pattern reversed. 
Despite the highest reported mental effort and cognitive load, ART produced the lowest recall performance (\textit{M} = 78.7).
WStE achieved the highest recall and lowest \textit{SD} (\textit{M} = 94.3, \textit{SD} = 12.7), indicating greater consistency, followed by GPT and WSiE (\textit{M} = 88.7 and 82.3).
ANOVA results indicated a statistically significant effect of stimulus type for both AIME and NASA TLX (all $p<0.001$), with ART differing significantly from all other conditions.
This means that the ART stimuli imposed substantially greater cognitive demands on listeners than any of the other stimulus types.
Additionally, ART recall was significantly lower than WStE ($p=0.019$) and numerically lower than GPT and WSiE, though these latter differences did not reach significance.
No statistically significant differences were observed between the WSiE, WStE, and GPT stimuli.

\begin{table}[t]                                                                       
\caption{Mean (SD) per stimulus type and omnibus test results. $^{a,b}$Groups sharing a superscript letter are not significantly different from each other ($\alpha = 0.05$). $^{*}p < 0.001$; $^{**}p = 0.019$. All \textit{F}-values Greenhouse--Geisser corrected (\textit{F}\textsubscript{GG}).}                      
\label{tab:perceptual}                                                               
\centering    
\small                                                                               
\begin{tabular}{lcccc|c}                                                             
  \toprule
   & ART & WSiE & WStE & GPT & \textit{F}\textsubscript{GG} \\
  \midrule
  \textbf{AIME} ↓&
    \makecell{62.9 \\(25.3)$^{a}$} &
    \makecell{28.9 \\(23.6)$^{b}$} &
    \makecell{28.7 \\(19.9)$^{b}$} &
    \makecell{25.2 \\(17.2)$^{b}$} &
    43.7$^{*}$ \\ \hline
  \textbf{TLX} ↓&
    \makecell{46.8 \\(24.0)$^{a}$} &
    \makecell{16.7 \\(16.5)$^{b}$} &
    \makecell{17.8 \\(16.0)$^{b}$} &
    \makecell{14.8 \\(14.3)$^{b}$} &
    58.6$^{*}$ \\ \hline
  \textbf{Recall} ↑&
    \makecell{78.7 \\(30.7)$^{a}$} &
    \makecell{82.3 \\(25.9)$^{ab}$} &
    \makecell{94.3 \\(12.7)$^{b}$} &
    \makecell{88.7 \\(21.2)$^{ab}$} &
    3.7$^{**}$ \\
  \bottomrule
\end{tabular}
\vspace{-0.5cm}
\end{table}

\section{Discussion}

\textbf{Conversational register and listenability.}
The results suggest that speech synthesised from written registers was substantially more listenable than its conversational counterpart.
ART, the source most closely resembling spoken language by register and classified as Informational interaction, imposed more than twice the mental effort and yielded the lowest comprehension scores of any stimulus type.
Hence, resemblance to spoken language (high D1) does not guarantee ease of auditory processing, and the register features that make speech engaging in face-to-face interaction impose a cognitive cost in a TTS context.


\noindent\textbf{Written register variation and listenability.}
Register differences across the Wikipedia-sourced and GPT-generated text did not translate into measurable listenability differences despite a readability gradient (WSiE most readable, WStE least). 
No statistically significant differences in mental effort, cognitive load, or cued recall were found among the three written sources.
The MDA profiles contextualise this result: although WStE (Learned exposition), WSiE, and GPT (General narrative exposition) occupy distinct registers, the three overlap in several parts of Biber's dimensional continuum.
This may be a sufficient condition for non-differentiable listenability under experimental conditions.
A notable exception is cued recall, where WStE achieved the highest scores (\textit{M} = 94.3), significantly exceeding ART despite imposing less listening effort.
This dissociation between cognitive load and retention — WStE imposes no more effort than simpler sources yet yields the best recall — warrants further investigation.
The listenability of GPT-generated text is also a notable finding.
Despite minimal prompting for listenability, GPT texts were statistically indistinguishable from both Wikipedia sources across all three measures and required the least mental effort of any source.
This suggests that LLM-generated text is a viable alternative to sourced written text for TTS delivery, with the practical advantage of on-demand generation and controllable register.
Finally, AIME and NASA TLX independently converged on the same pattern of results, supporting their concurrent validity as measures of listening effort and suggesting that either instrument may serve in future listenability studies.

\noindent\textbf{Limitations and future directions.}
Several limitations constrain these findings.
The stimulus set is small, and the use of a single TTS voice, uncontrolled online listening conditions, and the absence of controls for prior topic knowledge limit generalisability.
Furthermore, the one-way repeated-measures ANOVA does not account for topic as a factor, and a source-by-topic design would disentangle register effects from topic-specific difficulty.
A more fundamental limitation is that the four conditions do not share semantic content: topics are matched, but the actual information conveyed differs across sources, meaning content difficulty may co-vary with register as a confounding factor.
Future work could address this by generating written-register versions of the ART content, isolating register effects from semantic content.
Larger and more diverse stimulus sets, multiple TTS systems, and a wider range of listenability dimensions also remain to be explored.
A key open question is how to optimise the linguistic properties of text for verbal delivery, given that speech conveys information far beyond the textual signal.

\section{Conclusion}
This paper presents a preliminary study of listenability, comparing four TTS stimulus types drawn from distinct linguistic registers: talkback radio transcripts, Wikipedia Standard English, Wikipedia Simple English, and GPT-generated text.
Text characteristics were examined through text-based measures of readability and register, alongside a perceptual experiment measuring mental effort, cognitive load, and cued recall.
ART imposed significantly higher mental effort and cognitive load than all written sources, with no significant differences among WSiE, WStE, and GPT, indicating that register differences across written sources did not translate into measurable listenability differences.
WStE significantly outperformed ART in cued recall despite imposing comparable listening effort, and GPT-generated text performed on a par with both Wikipedia sources, supporting its viability as TTS input.
Speech synthesized from spoken transcripts proved the least listenable, suggesting that register features that make speech interactionally engaging in face-to-face interaction impose a cognitive cost in a TTS context.


\section{Acknowledgments}
Co-funded by the European Union under the Marie Skłodowska-Curie Grant Agreement No 101081465 (AUFRANDE). Views and opinions expressed are however those of the author(s) only and do not necessarily reflect those of the European Union or the Research Executive Agency, which cannot be held responsible for them.

\section{Generative AI Use Disclosure}
We acknowledge the use of generative AI in this work.
OpenAI's GPT-4o mini API was used to generate the GPT stimulus texts as part of the experimental design.
Claude (Anthropic) and ChatGPT (OpenAI) were used to assist in editing and polishing the manuscript.
All authors reviewed the full manuscript, take responsibility for its content, and consent to its submission.


\bibliographystyle{IEEEtran}
\bibliography{mybib}

\end{document}